\documentclass{article}

\usepackage{CJKutf8}

\usepackage{microtype}
\usepackage{graphicx}
\usepackage{subfigure}
\usepackage{booktabs} % for professional tables

\usepackage{hyperref}

\usepackage[accepted]{icml2021}

\icmltitlerunning{CLIPDraw: Exploring Text-to-Drawing Synthesis through Language-Image Encoders}

\begin{document}

\twocolumn[{
\icmltitle{CLIPDraw: Exploring Text-to-Drawing Synthesis\\through Language-Image Encoders}

% It is OKAY to include author information, even for blind
% submissions: the style file will automatically remove it for you
% unless you've provided the [accepted] option to the icml2021
% package.

% List of affiliations: The first argument should be a (short)
% identifier you will use later to specify author affiliations
% Academic affiliations should list Department, University, City, Region, Country
% Industry affiliations should list Company, City, Region, Country

% You can specify symbols, otherwise they are numbered in order.
% Ideally, you should not use this facility. Affiliations will be numbered
% in order of appearance and this is the preferred way.
\icmlsetsymbol{equal}{*}

\begin{center}
\textbf{Kevin Frans$^{1,2}$,  L.\ B.\ Soros$^1$ and  Olaf Witkowski$^{1,3,4}$}

% \mbox{}\\
$^1$Cross Labs, Cross Compass Ltd., Tokyo, Japan\\
$^2$Massachusetts Institute of Technology, Cambridge, MA, USA \\
$^3$Earth-Life Science Institute, Tokyo Institute of Technology, Japan\\
$^4$College of Arts and Sciences, University of Tokyo, Japan\\
kvfrans@csail.mit.edu
\end{center}

% kvfrans@csail.mit.edu}

% \author{Kevin Frans$^{1,2}$,  L.\ B.\ Soros$^2$ and  Olaf Witkowski$^{2,3,4}$ \\
% \mbox{}\\
% $^1$Massachusetts Institute of Technology, Cambridge, MA, USA \\
% $^2$Cross Labs, Cross Compass Ltd., Tokyo, Japan\\
% $^3$Earth-Life Science Institute, Tokyo Institute of Technology, Japan\\
% $^4$College of Arts and Sciences, University of Tokyo, Japan\\

% kvfrans@csail.mit.edu}

% \begin{icmlauthorlist}
% \icmlauthor{Kevin Frans}{mit, crosslabs}
% \icmlauthor{L.B. Soros}{crosslabs}
% \icmlauthor{Olaf Witkowski}{crosslabs, tt, todai}
% \end{icmlauthorlist}

% \icmlaffiliation{mit}{Massachusetts Institute of Technology, Cambridge, MA, USA}
% \icmlaffiliation{crosslabs}{Cross Labs, Cross Compass Ltd., Tokyo, Japan}
% \icmlaffiliation{tt}{Earth-Life Science Institute, Tokyo Institute of Technology, Japan}
% \icmlaffiliation{todai}{College of Arts and Sciences, University of Tokyo, Japan}

% \icmlcorrespondingauthor{Cieua Vvvvv}{c.vvvvv@googol.com}
% \icmlcorrespondingauthor{Eee Pppp}{ep@eden.co.uk}

\vskip 0.1in

\includegraphics[width=\textwidth]{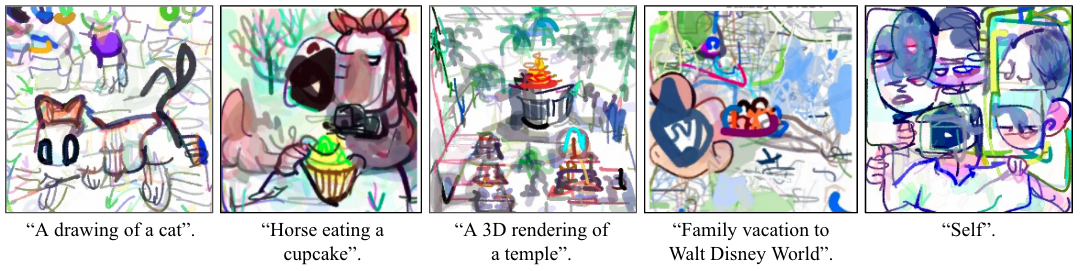}
\textbf{Various drawings synthesized by CLIPDraw}, along with the corresponding description prompts used. CLIPDraw synthesizes images from text by performing gradient descent over a set of RGBA Bézier curves, with the goal of minimizing cosine distance between the CLIP encodings of generated images and description prompts. CLIPDraw does not require learning a new model, and can generally synthesize images within a minute on a typical GPU.
\label{fig:mutations}
% \end{center}

\vskip 0.3in

}]

% \begin{figure}
%   \includegraphics[width=\linewidth]{imgs/header.png}
% \caption{Offspring produced by various genomes in Square Fitness world, according to their learned mutation behavior. 
% Population-based evolution genomes produce offspring in a ring corresponding to the nearest set of neighboring squares, allowing for efficient exploration. Genomes in the static-fitness environment learn to reduce the effect of mutation, whereas mutation functions created from single-genome evolution and random drift show no clear behavior.}
% \label{fig:mutations}
% \end{figure}

% this must go after the closing bracket ] following \twocolumn[ ...

% This command actually creates the footnote in the first column
% listing the affiliations and the copyright notice.
% The command takes one argument, which is text to display at the start of the footnote.
% The \icmlEqualContribution command is standard text for equal contribution.
% Remove it (just {}) if you do not need this facility.

%\printAffiliationsAndNotice{}  % leave blank if no need to mention equal contribution
% \printAffiliationsAndNotice{\icmlEqualContribution} % otherwise use the standard text.

\begin{abstract}
This work presents CLIPDraw, an algorithm that synthesizes novel drawings based on natural language input. CLIPDraw does not require any training; rather a pre-trained CLIP language-image encoder is used as a metric for maximizing similarity between the given description and a generated drawing. Crucially, CLIPDraw operates over vector strokes rather than pixel images, a constraint that biases drawings towards simpler human-recognizable shapes. Results compare between CLIPDraw and other synthesis-through-optimization methods, as well as highlight various interesting behaviors of CLIPDraw, such as satisfying ambiguous text in multiple ways, reliably producing drawings in diverse artistic styles, and scaling from simple to complex visual representations as stroke count is increased. Code for experimenting with the method is available at: \href{
https://colab.research.google.com/github/kvfrans/clipdraw/blob/main/clipdraw.ipynb}
{
\textbf{https://colab.research.google.com/github/\\kvfrans/clipdraw/blob/main/clipdraw.ipynb}
}
\end{abstract}

\section{Introduction}

As humans, when we hear a description of a scene, it’e easy to picture what it may look like in our heads. Conversely, when we construct a mental image, it’s easy to describe that scene in words. At some level, humans have a deeply coupled representation for both textual and visual structures that is key to understanding our everyday world.

The recent introduction of CLIP \citep{radford2021learning}, a dual language-image encoder, is a large step towards unifying textual and visual information. In a CLIP model, both text and images are mapped onto the same representational space, thus enabling the similarity between images and textual descriptions to be measured. When trained on large amounts of data, CLIP representations have been shown to solve a robust range of image-based recognition tasks.

This work presents \emph{CLIPDraw}, an algorithm that synthesizes novel drawings based on natural language input. CLIPDraw does not require any training; rather a pre-trained CLIP model is used as a metric for maximizing similarity between the given description and a generated drawing. Rather than photorealistic images, CLIPDraw aims to synthesize simple drawings that nevertheless match the prompt. Thus, CLIPDraw optimizes a set of vector strokes rather than pixel images, a constraint that biases drawings towards simple human-recognizable shapes.

The aim of this work is to present CLIPDraw a testbed for exploring language-image relationships and synthesizing AI-assisted artwork, as well as to showcase various nuances of the method. Results compare between CLIPDraw and other optimization-based text-to-image methods, along with highlighting several interesting behaviors:
\begin{itemize}
\item By adjusting descriptive adjectives, such as “watercolor” or “3D rendering”, CLIPDraw produces drawings of vastly different styles.

\item CLIPDraw often matches the description prompt in creative ways, such as writing words from the prompt inside the image itself, or interpreting ambiguous nouns in multiple ways.

\item Running CLIPDraw with a low stroke count results in cartoonish drawings, while high stroke counts tend to result in realistic renderings.

\item By giving CLIPDraw abstract prompts, such as “happiness” or “self”, we can examine what visual concepts the CLIP model associates with them.

\item CLIPDraw behavior can be further controlled through the use of negative prompts, such as “a messy drawing”, to encourage the opposite behavior.
\end{itemize}

\section{Related Work}

\begin{figure*}
  \includegraphics[width=\linewidth]{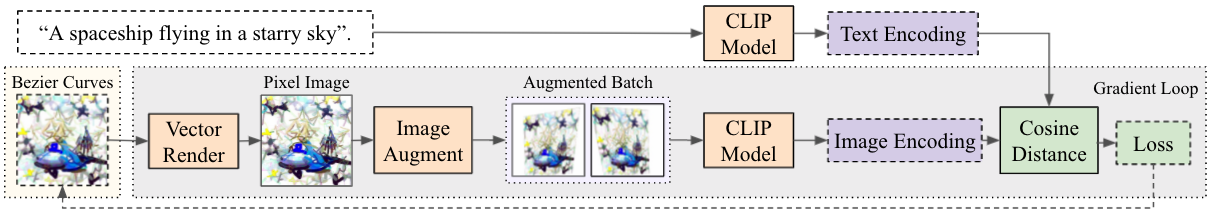}
\caption{\textbf{CLIPDraw iteratively synthesizes images through evaluation-time gradient descent.} Starting from a random set of Bézier curves, the position and colors of the curves are optimized so that the generated drawings best match the given description prompt. Before being passed into the CLIP encoder, drawings are augmented into multiple perspective-shifted copies.}
\label{fig:method}
\end{figure*}

\begin{figure*}
  \includegraphics[width=\linewidth]{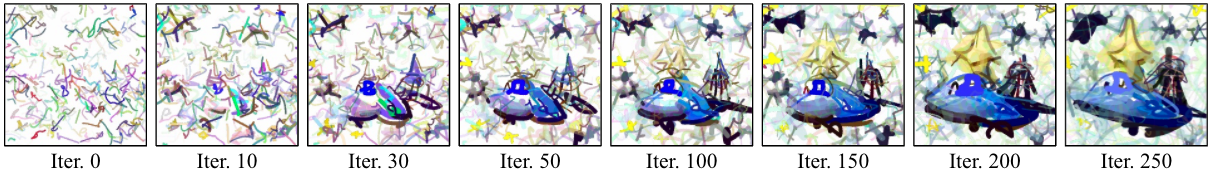}
\caption{\textbf{A typical CLIPDraw run gradually forms messy curves into concrete shapes.} In this example, the drawing first develops a background of star-shaped structures, which eventually develop into a large spaceship. Near the later iterations, more pronounced stars appear, in addition to a Darth Vader-like figure riding the spaceship.}
\label{fig:iter}
\end{figure*}

% \begin{figure*}
%   \includegraphics[width=\linewidth]{imgs/method.png}
% \caption{\textbf{Top: CLIPDraw iteratively synthesizes images through evaluation-time gradient descent.} Starting from a random set of Bézier curves, the position and colors of the curves are optimized so that the generated drawings best match the given description prompt. \textbf{Bottom: A typical CLIPDraw run gradually forms messy curves into concrete shapes.} In this example, the drawing first develops a background of star-shaped structures, which eventually develop into a large spaceship. Near the later iterations, more pronounced stars appear, in addition to a Darth Vader-like figure riding the spaceship.}
% \label{fig:method}
% \end{figure*}

\textbf{Text-to-Image Synthesis.} This work greatly draws from the field of text-to-image synthesis, whose primary aim is to generate images that correctly match a given textual description. In recent years, focus has been on methods that aim to learn a direct text-to-image mapping function, often in the form of a conditional GAN \citep{goodfellow2014generative, mirza2014conditional, reed2016generative, frolov2021adversarial}. Commonly used datasets include Oxford-120 Flowers \citep{nilsback2008automated}, CUB-200 Birds \citep{wah2011caltech}, and COCO \citep{lin2014microsoft}, all of which are comprised of natural images along with a set of captions describing them. While GAN-based methods have enabled considerable progress towards photorealistic image synthesis, strong autoregressive models have achieved similar quality results \citep{oord2017neural, chen2020generative}, with the recent DALL-E model \citep{ramesh2021zero} showcasing the benefit of scaling text-to-image synthesis networks to a large capacity. In comparison to text-to-image generative models, which require large amounts of training, this work follows the framework of \emph{synthesis through optimization}, in which images are generated through evaluation-time optimization against a given metric.

\textbf{Synthesis Through Optimization.} Instead of directly learning an image generation network, an alternative method of image synthesis is to optimize towards a matching image during evaluation time. This framework is often referred to as \emph{activation maximization} \citep{erhan2009visualizing, nguyen2016synthesizing, deepdream}, where a random image is optimized through backpropogation to increase certain neuron activations of a pretrained network. Activation-maximization methods have produced highly realistic images, however it is a challenge to understand the meaning of a neuron activation. CLIPDraw builds off a set of methods where rather than maximizing an activation, the objective is to minimize the distance between the produced image and a given description phrase, as defined by a powerful CLIP language-image encoder \cite{fernando2021generative, bigsleep, galatolo2021generating}. A key issue in synthesis through optimization is that the produced images often leave the space of natural images \citep{nguyen2015deep}, or fool the system through adversarial means \citep{goodfellow2014explaining}, thus a body of work aims to discover ‘natural image priors’ to constrain which images may be produced \citep{nguyen2016synthesizing, nguyen2017plug}. While a typical solution is to constrain optimization to the generative space of a GAN, this setup can be expensive to evaluate, and only allows synthesis of images producible by the GAN generator. Because CLIPDraw focuses on synthesizing drawings rather than realistic pictures, CLIPDraw instead limits optimization to a set of vector curves. This constraint results in stroke-based images, which must capture larger features such as shapes and outlines, rather than fine-grained textures.

\textbf{Vector Graphics.} This work builds largely from work by \citet{li2020differentiable}, which introduces a differentiable renderer for vector graphics. Image generation methods that operate over vector images have traditionally required a vector-based dataset, however recent work has shown how differentiable renderers can be used to bypass this limitation \citep{reddy2021im2vec, vec}. CLIPDraw uses a differentiable renderer as a representation for generating drawings; namely a set of RGBA Bézier curves are optimized rather than a matrix of pixels.

\section{Method}

The objective of CLIPDraw is to synthesize a drawing that matches a given description prompt (see Front Figure). Specifically, a pre-trained CLIP model is used as a judge. A CLIP model contains two networks -- an image encoder and a textual encoder -- which both map their respective inputs into a shared encoding space of a 512-length vector. Similarity is measured via the cosine distance between two encodings. Thus, the goal of CLIPDraw is to produce an image which, when encoded via CLIP, matches the CLIP encoding of the given description prompt.

Drawings in CLIPDraw are represented by a set of differentiable RGBA Bézier curves, following the method by \citet{li2020differentiable}. Each curve is parametrized by between 3 to 5 control points, along with thickness and an RGBA color vector. Drawings initially begin with curves randomly distributed throughout the image, with a white background as the default color. During optimization, the number of curves and control points is fixed, however the positions of the points along with the thickness and color vectors can be optimized via gradient descent.

The CLIPDraw algorithm (Algorithm \ref{alg:clipdraw}) works by running evaluation-time gradient descent, as shown in Figure \ref{fig:method}. First, the description phrase is encoded via the CLIP model, and a random set of $N$ Bézier curves are initialized. During each iteration, the curves are rendered to a pixel image via the differentiable renderer, and the resulting image is then duplicated $D$ times and augmented by a random perspective shift and random crop-and-resize. The resulting batch of augmented images is passed into the CLIP image encoder, and the cosine distances to the description phrase are summed to form the loss value. Because all operations are differentiable, gradient descent can be run through the entire loop, optimizing the parameters of the curves to decrease loss. This procedure is repeated $I$ times, until convergence.

The goal of the image augmentation is to force drawings to remain recognizable when viewed through various distortions. Without image augmentation, synthesis-through-optimization methods often result in adversarial images that fulfill the numerical objective but are unrecognizable to humans. This work specifically uses the \texttt{torch.transforms.RandomPerspective} and \texttt{torch.transforms.RandomResizedCrop} functions in sequence. Note that the specific details of the augmentation were not the focus of this work, thus a more robust augmentation function may exist and is left to future research.

Figure \ref{fig:iter} showcases the gradual synthesis of a typical CLIPDraw drawing. Note that while the optimization process is largely deterministic, there is randomness in the initial curves and image augmentations, thus multiple runs of CLIPDraw can result in different drawings.

\begin{algorithm}[tb]
   \caption{CLIPDraw}
   \label{alg:clipdraw}
\begin{algorithmic}
   \STATE {\bfseries Input:} Description Phrase $desc$; Iteration Count $I$; Curve Count $N$; Augment Size $D$; Pre-trained CLIP model.
   \STATE \textbf{Begin:}
   \STATE Encode Description Phrase. \emph{EncPhr = CLIP(desc)}
   \STATE Initialize Curves. \emph{Curves$_{0..N}$ = RandomCurve()}
   \FOR{$i=0$ {\bfseries to} $I$}
        \STATE Render Curves to Pixels. \emph{Pixels = DiffRender(Curves)}
        \STATE Augment the Image. \emph{AugBatch$_{0..D}$ = Augment(Pixels)}
        \STATE Encode Image. \emph{EncImg = CLIP(AugBatch)}
        \STATE Compute Loss. \emph{Loss = $-$CosineSim(EncPhr, EncImg)}
        \STATE Backprop. \emph{Curves $\leftarrow$ Minimize(Loss)}
        
   \ENDFOR
\end{algorithmic}
\end{algorithm}

\begin{figure*}
  \includegraphics[width=\linewidth]{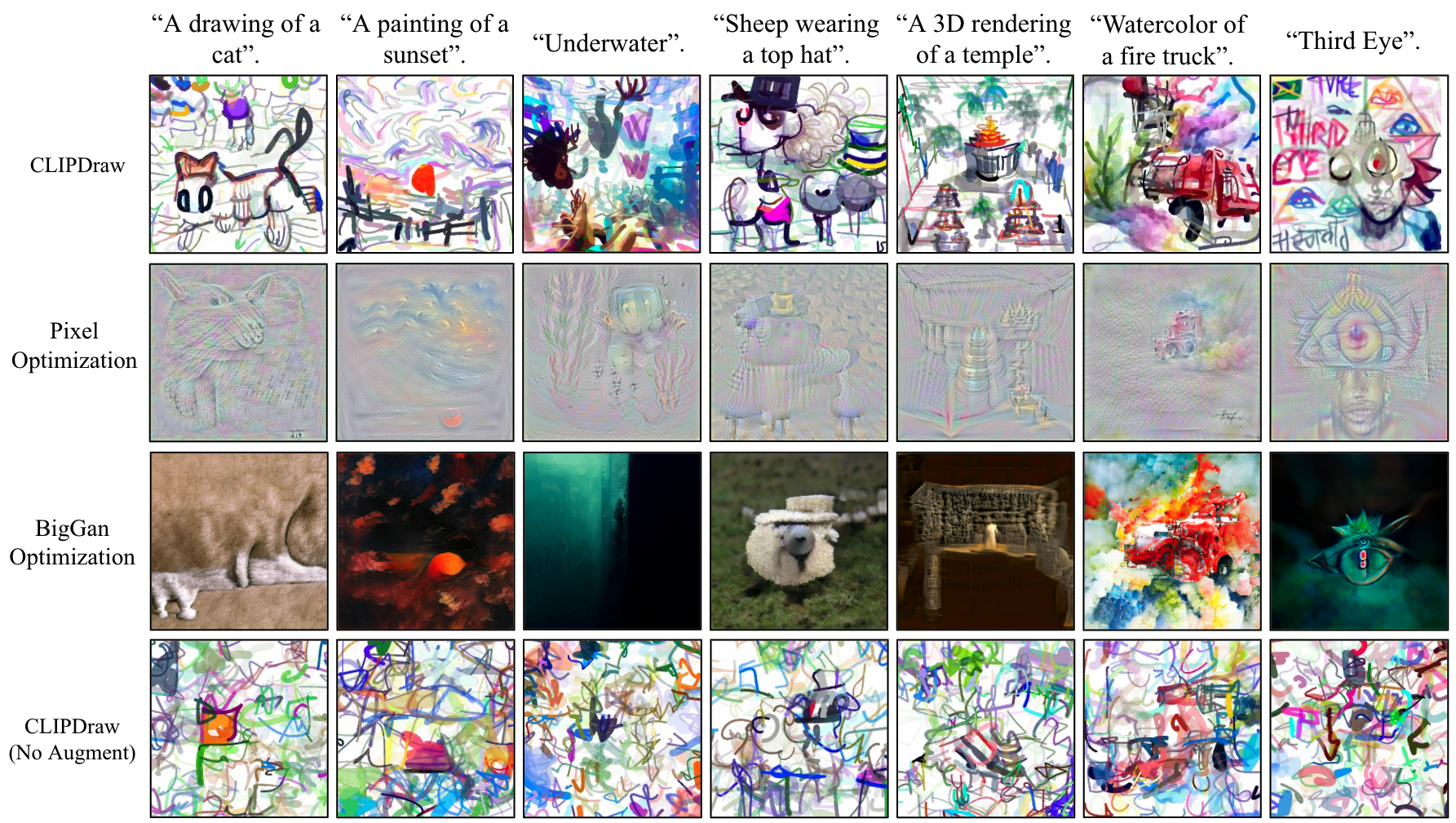}
  \caption{\textbf{Images synthesized via various synthesis-through-optimization methods}, with the goal of matching a given CLIP-encoded description phrase. CLIPDraw can produce a diverse set of human-recognizable drawings based on simple strokes and shapes.}
% \caption{\textbf{Images synthesized via various synthesis-through-optimization methods, with the goal of matching the CLIP encoding of a given description phrase.} CLIPDraw tends to result in a diverse set of human-recognizable drawings based on simple strokes and shapes. On the other hand, Pixel Optimization creates interesting textures but fails to compose colors and shapes. BigGan Optimization can synthesize high-resolution images, but is constrained to the set of images its generator can produce, thus it fails in out-of-distribution prompts. CLIPDraw without image augmentation produces images that score high numerically, but are nonsense when viewed by humans.}
\label{fig:comparison}
\end{figure*}

\section{Results}

In the following sections, various interesting behaviors of CLIPDraw are highlighted through a variety of examples. With the exception of Section 4.1, example images are picked to best convey the behavior in consideration. Focus is placed on qualitative observations, unusual behavior, or recurring trends in CLIPDraw image synthesis.

\subsection{How does CLIPDraw compare to other synthesis-through-optimization methods?}

Compared to methods that learn a direct generative model, optimization-based synthesis methods do not require prior training. Instead, images are generated through an evaluation-time optimization loop, aiming to maximize a given objective. This work specifically focuses on synthesizing images that match the CLIP encoding of a description prompt. The following methods are compared:

\begin{itemize}
\item \textbf{CLIPDraw}, in which drawings are produced by a set of RGBA Bézier curves. The control points, thickness, and colors of the curves can all be adjusted.

\item \textbf{Pixel Optimization}, which instead optimizes a 224x224x3 matrix of RGB pixel colors. Otherwise, all algorithmic aspects are the same as CLIPDraw, including image augmentation.

\item \textbf{BigGan Optimization}, in which images are produced using a pre-trained BigGAN generator. The weights of the generator are frozen; only the latent $Z$ vectors are optimized. Samples are generated using the method by \citet{bigsleep}.

\item \textbf{CLIPDraw (No Augment)}, which is identical to CLIPDraw, except no image augmentation is applied to the synthesized drawings.
\end{itemize}

In Figure \ref{fig:comparison}, various methods are run on the same CLIP matching objective. Each method is run for 250 steps of gradient descent. In the CLIPDraw results, stroke count is 256, and 8 duplicates are used during image augmentation. CLIPDraw tends to result in a diverse set of human-recognizable drawings based on simple strokes and shapes. On the other hand, Pixel Optimization creates interesting textures but fails to compose colors and shapes. BigGan Optimization can synthesize high-resolution images, but is constrained to the set of images its generator can produce, thus it fails in out-of-distribution prompts. CLIPDraw without image augmentation produces images that score high numerically, but are nonsense when viewed by humans.

\begin{figure}[H]
  \includegraphics[width=\linewidth]{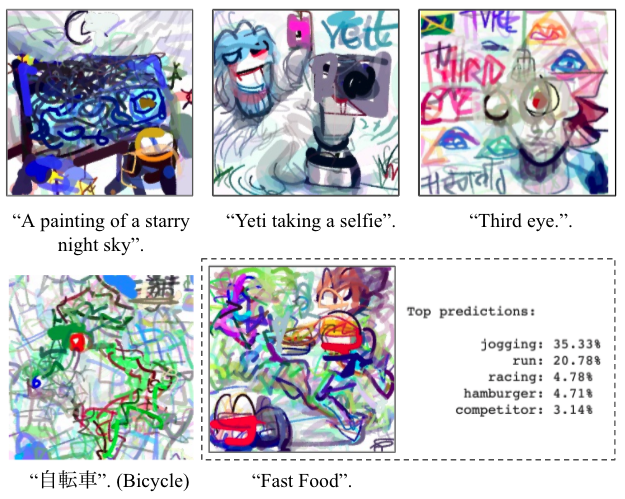}
\caption{\textbf{CLIPDraw often matches the description prompt through a variety of creative techniques}, such as forming letters inside the images, or interpreting ambiguous words in multiple ways.}
\label{fig:modes}
\end{figure}

\subsection{What kinds of visual techniques does CLIPDraw use to satisfy the textual description?}

CLIPDraw often results in drawings that match their description prompts in multiple, unexpected ways, as shown in Figure \ref{fig:modes}. A prime example is the prompt for “a painting of a starry night sky”. In this drawing, the main background features a sky with a prominent moon and a few scattered stars. The drawing itself is rendered in a painterly style, however the drawing also features an actual painting canvas and painter. Inside the canvas, black and blue swirls resemble Van Gogh’s 1889 “The Starry Night”. 

Another interesting behavior of CLIPDraw is its tendency to write words in the drawing itself. In “Yeti taking a selfie”, letters resembling “Yeti” can be seen in the top-right corner. In “Third Eye”, again words resembling “third” and “eye” are scattered throughout the image. 

\begin{CJK}{UTF8}{min}

At times, the drawings contain symbols that do not literally contain the description, but are tangentially associated, such as the prompt “自転車” (\emph{bicycle} in Japanese) resembling a Google Maps screenshot with a Japanese-like character in the corner.
\end{CJK}

The ambiguity of prompts also presents intriguing results. In the prompt “Fast Food”, a McDonald's logo along with a set of hamburgers is shown. However, also present are two joggers in a footrace, providing another interpretation of the phrase “fast”. Included in Figure \ref{fig:modes} are the top words predicted by CLIP as being closest to the image, showing that CLIP recognizes both “jogging” and “hamburger” as terms related to the synthesized drawing of “fast food”.

\begin{figure}[H]
  \includegraphics[width=\linewidth]{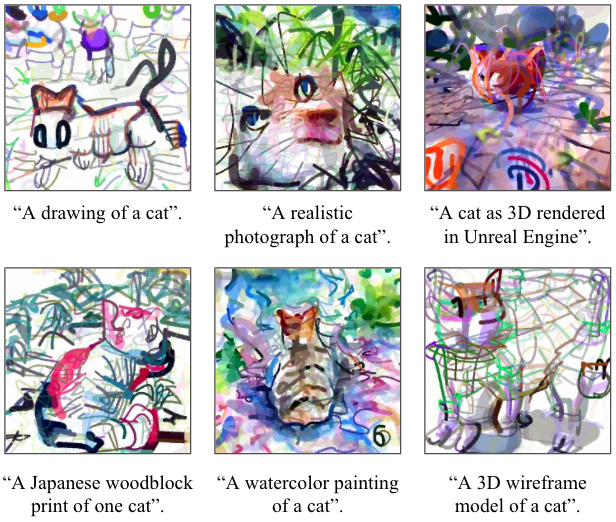}
\caption{\textbf{By adjusting descriptive adjectives, CLIPDraw can produce drawings of vastly different styles.} Styles vary not only in the texture of the images, but showcase different representations of the underlying content, such as a cartoonish cat when prompted for a "drawing", versus a cat in perspective when prompted for a "3D wireframe model".}
\label{fig:style}
\end{figure}

\subsection{Can CLIPDraw reliably produce drawings in different styles?}

A useful feature of CLIPDraw is its ability to adjust not just the content of its drawings, but also the styles, based on the description prompts given. Part of this flexibility is due to the robustness of curve-based images: in comparison to methods that use a pre-trained GAN generator, CLIPDraw drawings are not limited to the space of natural images. Thus, a variety of styles can be produced, and these styles are easily explorable through text.

As shown in Figure \ref{fig:style}, a synthesized image of a cat can look vastly different depending on the descriptor words included. When asked for a “drawing of a cat”, CLIPDraw synthesized a cartoonish depiction of a cat, comprised mostly of an outline and simple face. A “realistic photograph” features more detailed shading, while a “cat as 3D rendered in Unreal Engine” showcases complex lighting along with a depth-based blurring. Further styles feature a bias towards certain colors, such as the reds and greens of Japanese woodblock prints, or the multi-color blends of watercolors. 

An interesting result is that adjusting descriptive adjectives not only changes the textures of the drawings, akin to Style Transfer methods \citep{gatys2015neural}, but also changes its structural representation of the underlying content. For example, prompting for ``a drawing'' produces a flat cartoonish cat, while prompts like ``a 3D wireframe'' produce a cat in perspective, with depth and shadows.

\begin{figure}[H]
  \includegraphics[width=\linewidth]{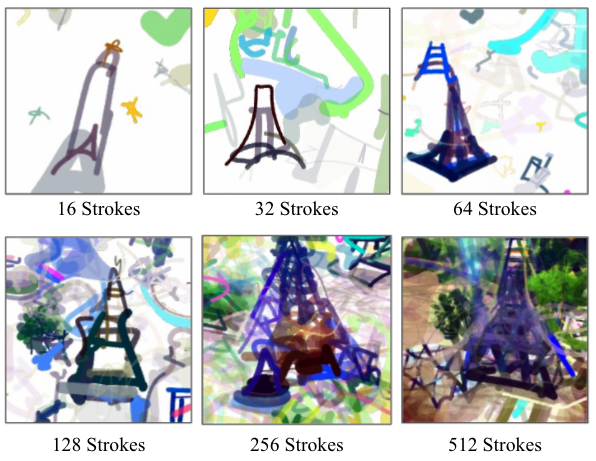}
\caption{\textbf{When stroke count is increased, CLIPDraw produces drawings of increasing realism.} Shown are multiple runs of the prompt ``The Eiffel Tower'' with a range of stroke counts. Low-stroke drawings opt for an abstract cartoonish representation, while high-stroke drawings capture 3D depth, background content, and complex shading.}
\label{fig:stroke}
\end{figure}

\subsection{How does the stroke count affect what drawings CLIPDraw produces?}

A crucial parameter of CLIPDraw is the number of strokes in each drawing. When the stroke count is low, CLIPDraw tends to produce cartoonish or abstract representations of the given description prompt. As stroke counts are increased, drawings become more detailed and incorporate additional features. Figure \ref{fig:stroke} showcases results for “The Eiffel Tower” on various stroke counts. In the 16-stroke example, the tower is drawn as only a few straight lines. The 32 and 64-stroke examples begin to display signs of 3D structure, such as a square base and a triangle-like scaffolding. Higher stroke count images begin more details on the Eiffel Tower itself, along with additional features such as background colors and complex lighting.

A common thread in synthesis-through-optimization methods is that pure optimization leads to undesirable results; it is also necessary to constrain optimization to a suitable space of images, such as the natural images generated by a GAN, or any image made of strokes in the case of CLIPDraw. Limiting stroke count furthers this constraint.  When optimizing within the space of 16-stroke images, it is hard to achieve details or textures, thus synthesized drawings will reveal the most basic forms that make up a visual concept. As a tool for AI-assisted art, the stroke-count parameter presents an easy way of adjusting between ``simple'' and ``complex''. This behavior may be useful in applications such as UI or icon design, where simplicity is important.

\begin{figure}[H]
  \includegraphics[width=\linewidth]{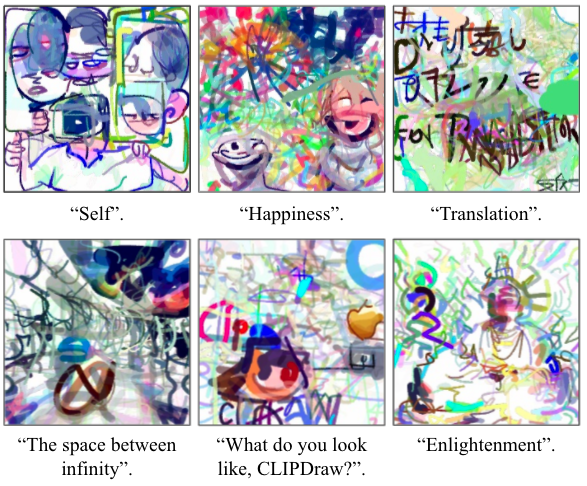}
\caption{\textbf{Abstract prompts grant insight into how CLIP relates visual concepts.} Synthesized images often contain symbols that indirectly relate to the description prompt through a cultural connection.}
\label{fig:abstract}
\end{figure}

\subsection{What happens if abstract words are given as a description prompt?}

When given an abstract prompt without a literal interpretation, CLIPDraw must utilize cultural connections to come up with visual concepts that relate to the description. Often, this results in drawings that contain symbols relating to the given prompt, such as in Figure \ref{fig:abstract} with “Happiness” containing smiling faces and fireworks, “Translation” showcasing English and Japanese-like characters, and “Enlightenment” featuring a prominent monk-like figure.

At times, synthesized drawings demonstrate concepts through more complex relationships. In the prompt “Self”, the resulting drawing features a body with multiple heads, evoking e.g. the idea that a person’s self may contain multiple outward personalities. When asked “What do you look like, CLIPDraw?”, the synthesized drawing contains a smiling face followed by text resembling “CLIPDRAW”. Finally, “The space between infinity” presents a dream-like landscape with an infinity symbol under a galaxy-filled sky.

The ability for CLIPDraw to answer abstract prompts through related concepts presents a potential tool for exploration of human culture. As CLIP was trained on vast amounts of human data, cultural connections featured in synthesized drawings can grant insight into the typical connections humans may make. As a tool for AI-assisted art, this ability is also useful, as artists who want to evoke certain emotions will employ concepts that are culturally related to that emotion, akin to the smiles and fireworks in the drawing of “Happiness” above.

\begin{figure}[H]
  \includegraphics[width=\linewidth]{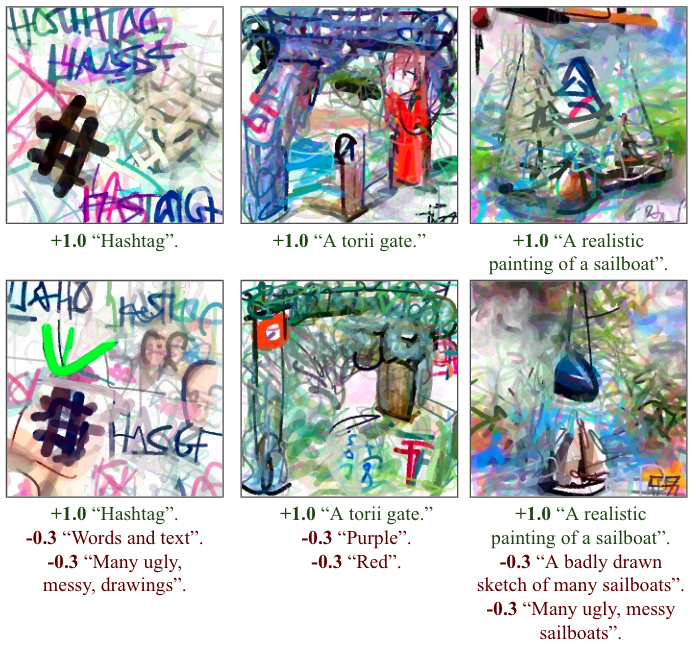}
\caption{\textbf{CLIPDraw behavior can be adjusted through negative description prompts.} Negative prompts discourage synthesized drawings from matching with them, presenting a tool for fine-tuning results.}
\label{fig:negative}
\end{figure}

\subsection{Can synthesized drawings be fine-tuned via additional negative prompts?}

A common pain point in AI-assisted image synthesis is that is hard to control what the AI will produce. One potential solution in CLIP-based methods is to introduce negative prompts. In this setup, the optimization objective is to minimize cosine distance between the CLIP-encoded drawing and the description prompt, while maximizing distance between the drawing and a set of negative prompts.

Figure \ref{fig:negative} showcases how negative prompts can fine-tune the behavior of CLIPDraw. Presented are pairs of drawings synthesized from the same random initialization, with the bottom row utilizing additional negative prompts, weighted on a 0.3:1 scale. In the “Hashtag” example, the original drawing contains many instances of the word “hashtag” written out. By penalizing “Words and text”, the bottom example contains fewer words, instead featuring a set of faces typically seen in social media. A drawing of “a torii gate” originally results in a purple and red structure, however by penalizing “Purple” and “Red” the main color of the drawing switches to green. Lastly, the original drawing for “a realistic painting of a sailboat” features many sailboats on an ocean, and penalizing the phrase “many sailboats” results in a drawing featuring only a singular sailboat in the center.

In general, while negative prompts present a richly semantic way to fine-tune image synthesis in CLIP-based methods, it remains tricky to locate prompts that consistently encourage the intended behavior. Many times, negative prompts show negligible effect on the resulting drawing. During experiments, a cure-all negative prompt such as “a low-quality drawing”, with the goal of consistently improving drawing quality, was unable to be found. Further work remains on how to best influence CLIP-based synthesis-through-optimization methods through additional objectives, whether negative or positive.

\section{Discussion}

This work presents CLIPDraw, a text-to-drawing synthesis method based on the CLIP language-image encoder. CLIPDraw does not require any model training; rather, drawings are synthesized through iterative optimization during evaluation time. CLIPDraw is not the first method to utilize evaluation-time optimization for image synthesis; in fact, many recent works have used CLIP as an objective as well. However, by constraining image synthesis to images made up of RGBA Bézier curves, CLIPDraw biases towards simple drawings of human-recognizable concepts. The focus of this paper is to examine the nuances of CLIPDraw behavior, and experiments focus on specific questions and observations about synthesized drawings.

\subsection{Limitations}

The CLIPDraw method presented comes with various limitations. First, CLIPDraw inherently biases towards drawings rather than photorealistic images. Thus, synthesizing high-resolution images is a challenge, and CLIPDraw will often fall short of methods that incorporate a high-functioning generative model. This problem is related to a classic pitfall in synthesis-through-optimization methods, which is that the numerical objective used is not necessarily what is desired, e.g. an image may very closely match the CLIP objective, while looking messy and ugly to a human. Thus it is important to introduce auxiliary objectives or constraints. In the case of CLIPDraw, drawings are constrained by the Bézier curve representation, however stricter constraints such as fooling a GAN discriminator may improve synthesis quality.

A second limitation has to do with using CLIP encodings as a synthesis objective. While CLIP provides a rich textual representation for describing an image, in comparison to coarser neuron-activation objectives, it still remains a challenge to specify details. For example, it is hard to tell CLIPDraw to move a sailboat to the other side of the image. This work explored negative prompts as a possible direction towards more fine-grained adjustments, however a consistently satisfying method was hard to locate. A promising path in future research can lie in how to correctly adjust synthesized images, or introduce finer detail into description prompts via additional objectives.

\subsection{Ethics and Social Biases}

An important concept to keep in mind when dealing with human data is the existence of inherent social biases contained within. The pre-trained CLIP model is trained on a large corpus of online data, so its representations may include connections or biases that are undesirable. As CLIPDraw does not learn a new model, but instead optimizes based on CLIP itself, the bias studies presented in the CLIP paper \citep{radford2021learning} are highly relevant for CLIPDraw as well. This work does not specifically present additional social bias studies, however it is important to keep in mind that these biases exist when applying CLIPDraw to real-world use cases. As an example, in Section 4.5, a use case for CLIPDraw is mentioned as a tool for exploring visual connections in human culture. It is crucial to recognize that symbols or connections that are formed by CLIPDraw are not necessarily reflective of human culture, but rather are artifacts of the data used to train the original CLIP model. Thus, while CLIPDraw can be used to synthesize drawings that utilize cultural connections to evoke emotions or abstract concepts, it remains the duty of the user to ensure that the final product is up to desired standards. This caveat is especially key in automated setups, where AI-assisted art may produce unwanted results when run without a human in the loop.

\subsection{Future Work}

Overall, the aim of this work is to introduce CLIPDraw as an easily accessible starting point to experiment with natural language image synthesis. Due to its focus on drawings rather than photorealistic rendering, CLIPDraw presents a straightforward method to examine language-image relationships without the overhead of realism. The presented CLIPDraw implementation can generally synthesize images within a minute on a typical GPU. Thus, interactable source code for experimenting the methods is made available at this Colab notebook: \href{
https://colab.research.google.com/github/kvfrans/clipdraw/blob/main/clipdraw.ipynb}
{
\textbf{https://colab.research.google.com/github/\\kvfrans/clipdraw/blob/main/clipdraw.ipynb}}

The results presented in this paper aim to describe various interesting behaviors of CLIPDraw, however they are by no means exhaustive. CLIP-based text-to-image synthesis remains a field with many promising directions, and we hope others will use this work as a backboard for additional research into the nuances of synthesis-through-optimization, or as a practical tool for AI-assisted art and other interactive visual applications.

\bibliography{main}
\bibliographystyle{icml2021}

%%%%%%%%%%%%%%%%%%%%%%%%%%%%%%%%%%%%%%%%%%%%%%%%%%%%%%%%%%%%%%%%%%%%%%%%%%%%%%%
%%%%%%%%%%%%%%%%%%%%%%%%%%%%%%%%%%%%%%%%%%%%%%%%%%%%%%%%%%%%%%%%%%%%%%%%%%%%%%%
% DELETE THIS PART. DO NOT PLACE CONTENT AFTER THE REFERENCES!
%%%%%%%%%%%%%%%%%%%%%%%%%%%%%%%%%%%%%%%%%%%%%%%%%%%%%%%%%%%%%%%%%%%%%%%%%%%%%%%
%%%%%%%%%%%%%%%%%%%%%%%%%%%%%%%%%%%%%%%%%%%%%%%%%%%%%%%%%%%%%%%%%%%%%%%%%%%%%%%

\end{document}